\documentclass[conference]{IEEEtran}
\IEEEoverridecommandlockouts
\usepackage{cite}
\usepackage{amsmath,amssymb,amsfonts}
\usepackage{algorithmic}
\usepackage{graphicx}
\usepackage{textcomp}
\usepackage{xcolor}
\def\BibTeX{{\rm B\kern-.05em{\sc i\kern-.025em b}\kern-.08em
    T\kern-.1667em\lower.7ex\hbox{E}\kern-.125emX}}
\begin{document}

\title{LLM-Grounded Explainable AI for Supply Chain Risk Early Warning 
via Temporal Graph Attention Networks}

\author{\IEEEauthorblockN{1\textsuperscript{st} Zhiming Xue*}
\IEEEauthorblockA{\textit{College of Engineering} \\
\textit{Northeastern University}\\
Boston, USA \\
xue.zh@northeastern.edu}
\and
\IEEEauthorblockN{2\textsuperscript{nd} Yujue Wang}
\IEEEauthorblockA{\textit{Department of Chemistry and Chemical Biology} \\
\textit{University of New Mexico}\\
Albuquerque, USA \\
yujue@unm.edu}
\and
\IEEEauthorblockN{3\textsuperscript{rd} Menghao Huo}
\IEEEauthorblockA{\textit{School of Engineering} \\
\textit{Santa Clara University}\\
Santa Clara, USA \\
menghao.huo@alumni.scu.edu}
}

\maketitle

\begin{abstract}
Disruptions at critical logistics nodes pose severe risks to global supply chains, yet existing risk prediction systems typically prioritize forecasting accuracy without providing operationally interpretable early warnings. This paper proposes an evidence-grounded framework that jointly performs supply chain bottleneck prediction and faithful natural-language risk explanation by coupling a Temporal Graph Attention Network (TGAT) with a structured large language model (LLM) reasoning module. Using maritime hubs as a representative case study for global supply chain nodes, daily spatial graphs are constructed from Automatic Identification System (AIS) broadcasts, where inter-node interactions are modeled through attention-based message passing. The TGAT predictor captures spatiotemporal risk dynamics, while model-internal evidence---including feature z-scores and attention-derived neighbor influence---is transformed into structured prompts that constrain LLM reasoning to verifiable model outputs. To evaluate explanatory reliability, we introduce a directional-consistency validation protocol that quantitatively measures agreement between generated risk narratives and underlying statistical evidence. Experiments on six months of real-world logistics data demonstrate that the proposed framework outperforms baseline models, achieving a test AUC of 0.761, AP of 0.344, and recall of 0.504 under a strict chronological split while producing early warning explanations with 99.6\% directional consistency. Results show that grounding LLM generation in graph-model evidence enables interpretable and auditable risk reporting without sacrificing predictive performance. The framework provides a practical pathway toward operationally deployable explainable AI for supply chain risk early warning and resilience management.
\end{abstract}


\section{Introduction}

Global supply chains are highly complex networks where disruptions at critical logistics nodes can cause cascading failures across the entire system. Maritime transport accounts for over 80
making port operations a critical component of modern supply
chain networks. Maritime hubs, such as the Port of Los Angeles and Long Beach, serve as representative bottlenecks in these networks, facilitating a massive portion of international trade \cite{b1}.  The systemic fragility of such supply chain nodes was underscored during recent global disruptions, where localized backlogs triggered worldwide delays and severe inventory shortfalls \cite{b2}. Because risks propagate rapidly from primary logistics hubs to inland freight networks and downstream manufacturing, providing supply chain managers and stakeholders with reliable, early predictive warnings is essential for proactive resource reallocation and risk mitigation.

Predicting supply chain bottlenecks and localized risks is a fundamental spatiotemporal challenge. Traffic density and logistics throughput are governed by localized conditions and complex spatial interactions across the network. While classical time-series models like ARIMA or LSTM treat logistics nodes as independent entities, Graph Neural Networks (GNNs) naturally encode spatial dependencies by modeling locations as nodes connected through relational edges. Among these models, Graph Attention Networks (GATs) \cite{b3} allow each node to adaptively weight neighbor influence; their learnable attention coefficients provide a mechanism to weight neighbor contributions, offering a mathematical pathway to audit which spatial nodes most heavily drive a specific risk forecast.

However, a significant "explainability gap" persists in supply chain AI literature. Most current research prioritizes aggregate performance metrics (e.g., AUC, $F_1$) over actionable, node-level insights required by non-technical supply chain managers. While post-hoc methods like GNNExplainer \cite{b4} or SHAP \cite{b5} identify influential subgraphs, their outputs remain numerical and abstract. Recent advances in Large Language Models (LLMs) for decision support offer a solution: by synthesizing structured evidence---such as feature z-scores and attention weights---LLMs can generate coherent, domain-specific early warning reports. To our knowledge, this work is the first to integrate attention-based spatial prediction with LLM-driven natural language explanations for supply chain risk early warning, while also addressing broader cost-performance considerations for managing LLM context length in deployed reasoning pipelines \cite{b6}.

This paper provides the following contributions:
\begin{enumerate}
    \item \textbf{Logistics Risk Dataset and Pipeline:} Using maritime AIS data as a representative supply chain case study, we develop a pipeline that transforms raw logistics broadcasts into a chronological sequence of 89 spatiotemporal graph snapshots, totaling $3.02 \times 10^{4}$ labeled node-day samples.
    \item \textbf{TGAT Risk Predictor:} We implement a Temporal Graph Attention (TGAT) model for bottleneck prediction. Our primary model achieves a test AUC of 0.761 on a highly imbalanced dataset (13.5\% positive rate) using a strict chronological split to prevent data leakage.
    \item \textbf{Attention-Based Evidence Extraction:} We bridge the gap between model internals and human interpretation by extracting per-node evidence records, including top-ranking features by z-score and primary spatial neighbors via attention-proxy weights.
    \item \textbf{LLM-Generated Risk Synthesis:} We demonstrate that GPT-4o-mini, guided by structured evidence prompts, generates multi-section early warning reports with 99.6\% directional consistency, providing a blueprint for transparent, AI-assisted supply chain decision support.
\end{enumerate}

\section{Related Work}

\subsection{Supply Chain Risk Prediction and Graph Modeling}

Graph-based approaches demonstrate clear advantages in modeling complex logistics and supply chain networks:
\cite{b7} shows spatio-temporal graph convolution improves throughput prediction at critical supply chain nodes over univariate baselines, while \cite{b8}
predicts bottleneck risks using spatially-derived features from logistics data. These works, however, primarily report aggregate metrics and provide
limited node-level interpretability for supply chain decision makers.  Graph Attention Networks
(GATs)~\cite{b3} extend GNNs with learnable neighbour
weights, yet in logistics applications attention signals are typically
used solely to boost predictive performance rather than to drive
structured downstream explanation.

\subsection{Explainable Graph Learning and LLM-based Reasoning}

As GNNs are applied in operational supply chain systems, the need for interpretable prediction has intensified. Post-hoc methods such as GNNExplainer~\cite{b4} and SHAP~\cite{b5} provide numerical attribution scores that often require expert interpretation. Recent research integrates LLMs into logistics decision support, demonstrating that structured quantitative signals can be synthesised into coherent natural-language analysis~\cite{b9}, and lightweight language models can support reliable agentic decision processes under constrained settings~\cite{b10}; context-efficient adaptation techniques that combine domain-specific fine-tuning with model ensembling have further been shown to enhance the reliability of such lightweight models on specialized reasoning tasks \cite{b11}. 

Nevertheless, existing systems typically act as descriptive summarisers without explicitly constraining generation to model-internal graph evidence, leaving faithfulness concerns unaddressed. The present work bridges these gaps by injecting attention weights and feature-level statistics as structured evidence records into a constrained LLM reasoning module, enabling both forecasting accuracy and auditable natural-language risk reporting within a unified pipeline.

\section{Data and Graph Construction}

\subsection{Spatiotemporal Graph Construction}

We construct daily graph snapshots from Automatic Identification System (AIS) broadcasts obtained from the NOAA Marine Cadastre archive (January--June 2023, UTM Zone 10, US West Coast). The study region spans $32^\circ\mathrm{N}$--$35^\circ\mathrm{N}$ latitude and $121^\circ\mathrm{W}$--$117^\circ\mathrm{W}$ longitude, covering the San Pedro Bay anchorage and the Port of Los Angeles/Long Beach complex.

The region is discretised into a regular $0.1^\circ \times 0.1^\circ$ grid ($\approx 11\,\mathrm{km} \times 11\,\mathrm{km}$ per cell at this latitude). For each day $t$, grid cells with at least one AIS observation form the active node set $V_t$. Nodes are connected via a fixed-topology $k$-nearest-neighbour graph with $k=8$ based on Euclidean distance between cell centroids, yielding a snapshot
$G_t = (V_t, E_t, X_t)$, where $X_t \in \mathbb{R}^{|V_t|\times 10}$.
Days with fewer than ten active nodes are discarded, resulting in 89 valid daily snapshots and approximately $3.02\times 10^4$ node--day samples.

Fig. 1 shows the spatial coverage of the study area with representative logistics activity density overlaid on the hub map. Each colored cell corresponds to an active graph node representing a supply chain sub-region; edge opacity encodes relative attention weight learned by the model.

\begin{figure}[htbp]
  \centering
  \includegraphics[width=0.75\columnwidth]{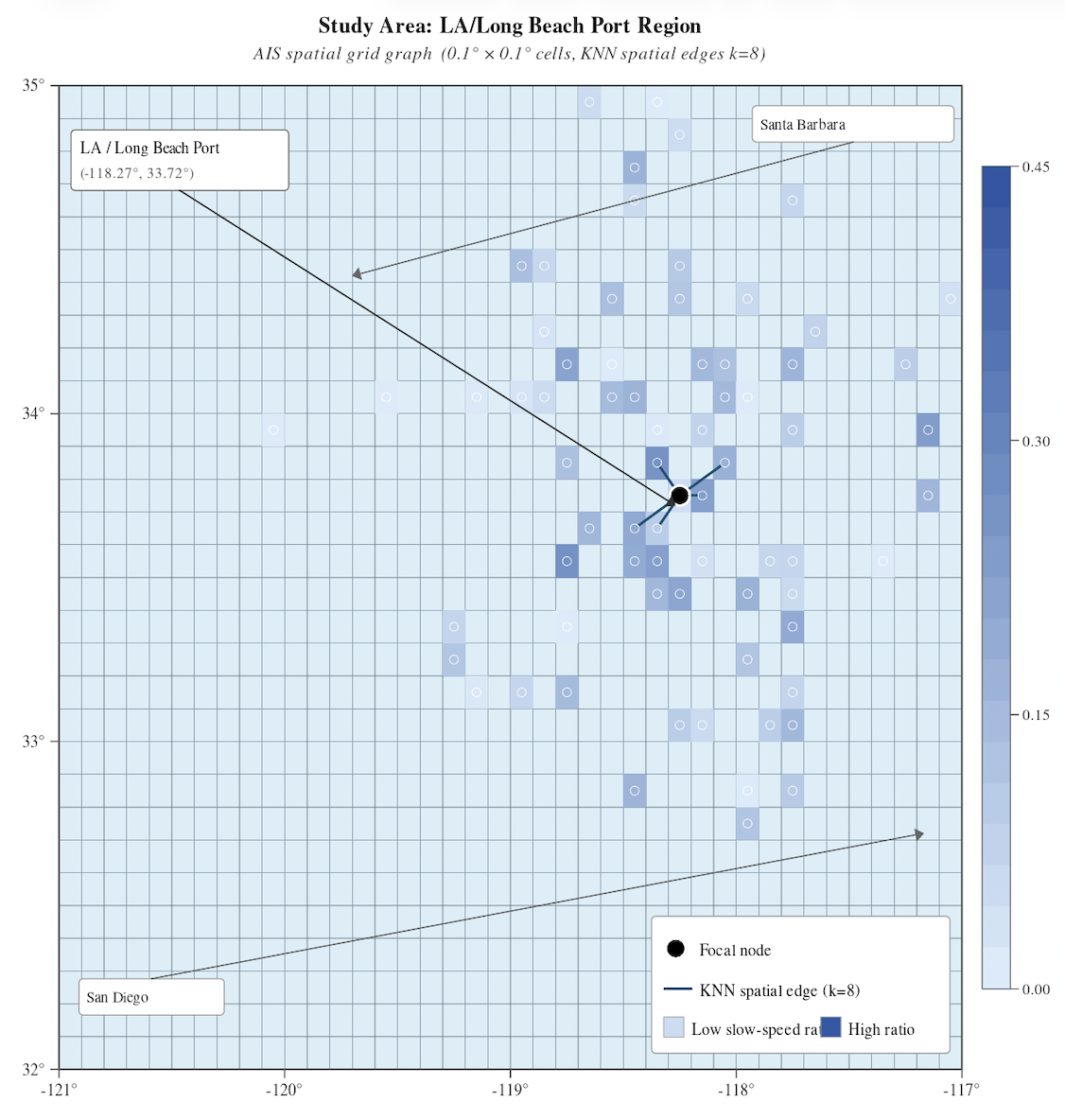}
  \caption{LA/Long Beach logistics hub study area and spatial graph overview}
  \label{fig:system_architecture}
\end{figure}

\subsection{Node Representation and Risk Escalation Label}

For each node--day pair, we compute ten kinematic and traffic-density features aggregated from AIS broadcasts within the cell boundary, including mean and standard deviation of speed over ground (SOG), slow-vessel ratio (SOG $<2$ knots), anchor ratio, vessel count, cargo and tanker ratios, mean vessel length and draft, and circular variance of course over ground (COG). These features collectively characterize the operational state of each supply chain node. All features are snapshot-wise $z$-score normalised across active nodes.

We define a binary supply chain risk escalation label as:
\[
y_i^{t} =
\begin{cases}
1, & \text{if } \texttt{slow\_ratio}_{i}^{t+1} > \texttt{slow\_ratio}_{i}^{t},\\
0, & \text{otherwise.}
\end{cases}
\]

The positive rate is approximately 13.5\%, indicating moderate class imbalance. 
Fig. 2 illustrates the point-biserial correlations between these ten features and the escalation label across the full dataset. Mean SOG (r=-0.204) and slow ratio (r=+0.190) exhibit the strongest associations with escalation, consistent with the physical intuition that cells already experiencing low-speed traffic are predisposed to further congestion. Conversely, the tanker ratio shows the weakest correlation (r=-0.017), reflecting the irregular scheduling of liquid bulk carriers at this complex. These empirical correlation directions are subsequently injected into the LLM prompt as ground-truth constraints, ensuring that the generated risk narratives are statistically grounded in the model's evidence.

\begin{figure}[htbp]
  \centering
  \includegraphics[width=1\columnwidth]{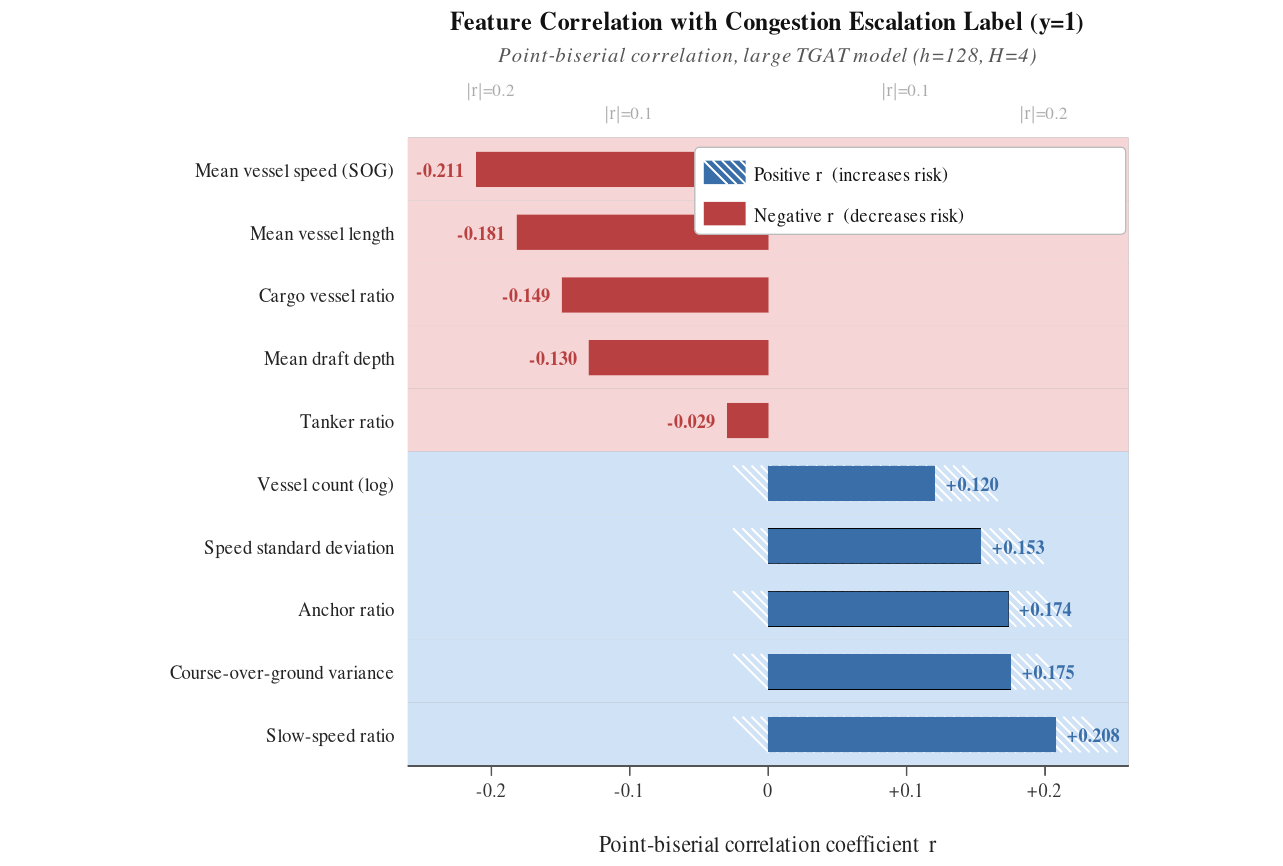}
  \caption{Feature-correlation chart}
  \label{fig:system_architecture}
\end{figure}

\section{Model Architecture and Training}

\subsection{Temporal Graph Attention Network}
To address the class imbalance (13.5\% positive rate) defined in Section III-B, we apply a positive-class weight of $6.74$ in the binary cross-entropy loss. The overall framework processes the ordered sequence of daily graph snapshots $G_1, G_2, \dots, G_{89}$ through three stages: spatiotemporal message passing, node-level binary classification, and attention-evidence extraction. Fig.3 summarizes the end-to-end pipeline.

\begin{figure}[htbp]
  \centering
  \includegraphics[width=0.6\columnwidth]{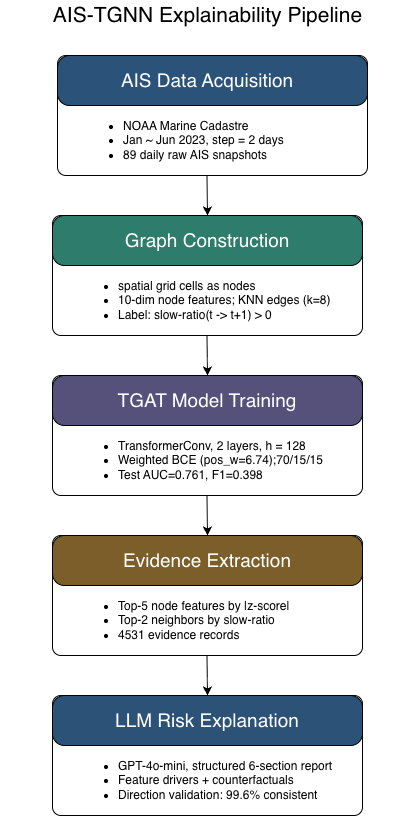}
  \caption{End-to-end pipeline architecture }
  \label{fig:system_architecture}
\end{figure}

At each time step $t$, the model receives the current graph snapshot $G_t = (V_t, E_t, \mathbf{X}_t)$, where $V_t$ is the set of active nodes, $E_t$ is the KNN edge set, and $\mathbf{X}_t \in \mathbb{R}^{|V_t| \times 10}$ is the node feature matrix. Each node $i$ aggregates information from its neighbors using a multi-head graph attention layer \cite{b3}. For head $m$, the attention coefficient $\alpha_{ij}^{(m)}$ from neighbor $j$ to node $i$ is computed as:

\begin{equation}
\alpha_{ij}^{(m)} = \text{softmax}_j \left( \text{LeakyReLU} \left( \mathbf{a}^{T} [ \mathbf{W} \mathbf{h}_i \parallel \mathbf{W} \mathbf{h}_j ] \right) \right)
\end{equation}

where $\mathbf{W} \in \mathbb{R}^{h \times 10}$ is a shared learnable weight matrix, $\mathbf{h}_i$ and $\mathbf{h}_j$ are the current feature vectors of nodes $i$ and $j$, $\parallel$ denotes concatenation, and $\mathbf{a}$ is a learnable attention vector. The aggregated node representation is the concatenation of $H$ independent attention heads, followed by a two-layer MLP with ReLU activation and dropout (rate 0.1) that maps each node's representation to a scalar risk-escalation logit. A sigmoid activation converts the logit to a predicted probability $p_i$.

Temporal context is incorporated by processing snapshots sequentially: the node embeddings produced at time step $t$ are used as prior-state features concatenated with the raw feature vector at time step $t+1$ before the next message-passing iteration, allowing the model to condition its predictions on recent traffic history without requiring a separate recurrent component. We train TGAT with hidden dimension $h = 128$ and $H = 4$ attention
heads, selected as the best-performing configuration from preliminary
experiments. All other hyperparameters are: dropout $= 0.1$,
learning rate $= 10^{-3}$, weight decay $= 10^{-4}$, Adam optimiser.

\subsection{Training Protocol}
The 89 daily snapshots are split chronologically: $70\%$ for training, $15\%$ for validation, and $15\%$ for testing. This strict chronological assignment prevents temporal data leakage. We train with binary cross-entropy loss weighted by the inverse positive-class frequency ($\text{pos\_weight} = 6.74$) to handle class imbalance. The model is trained for up to 50 epochs; the checkpoint achieving the highest validation AUC is retained for test evaluation. The decision threshold $\tau^*$ is chosen on the validation set to maximise $F_1$.

\begin{figure}[htbp]
  \centering
  \includegraphics[width=1\columnwidth]{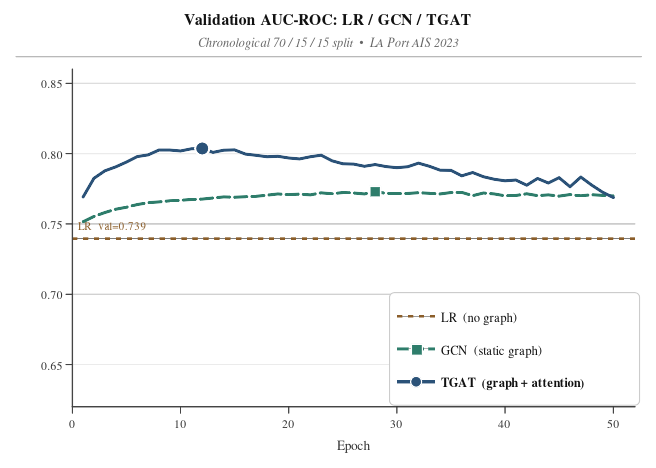}
  \caption{Validation AUC-ROC across training epochs. LR, GCN and TGAT}
  \label{fig:system_architecture}
\end{figure}

Fig.4 shows the validation AUC-ROC curves for all
three models across training epochs. LR provides a static reference
(val AUC $= 0.739$) with no iterative training. GCN improves
progressively but plateaus at approximately $0.773$, reflecting the
limitation of static graph convolution without attention. TGAT
converges to its peak at epoch~12 (val AUC $= 0.804$) and maintains
a consistent advantage over both baselines throughout training,
confirming that the attention mechanism captures spatiotemporal
patterns that neither LR nor GCN can represent. All models use
best-checkpoint selection; the checkpoint with the highest
validation AUC is retained for test evaluation.

\subsection{Attention Evidence Extraction}
After training, we record the predicted probability $p_i$ and raw attention logits $\alpha_{ij}$ for each node $i$. We sum the multi-head logits and apply softmax to obtain a scalar attention-proxy weight $w_{ij}$. For the LLM explanation stage, we retain an ``evidence record'' for each node comprising: the top-5 features ranked by $|z\text{-score}|$ with their correlation directions, and the top-2 spatial neighbours by $w_{ij}$ along with their most prominent features.

\section{LLM Explanation Module}

\subsection{Structured Prompt Design}
The LLM explanation module takes the evidence record for a single node--day as input and produces a structured natural-language risk report. We use GPT-4o-mini via the OpenAI API with a maximum output token budget of 2,500 and a fixed system prompt that instructs the model to act as a supply chain risk analyst and to base all reasoning exclusively on the provided structured evidence, without invoking external domain knowledge. This bounded, structured prompting design ensures computational feasibility. The user-turn prompt is assembled from four components: (1) a header block specifying the target grid cell, date, and predicted risk-escalation probability; (2) the top-5 feature table, listing each feature's $z$-score value, its correlation direction (\textit{increase risk} or \textit{decrease risk}), and the underlying point-biserial $r$; (3) the top-2 neighbour block, listing each neighbour's cell identifier, attention-proxy weight, and its single most prominent feature; and (4) a JSON response schema specifying the exact six output sections the model must populate.
\begin{figure*}[htbp]
  \centering
  \includegraphics[width=1.8\columnwidth]{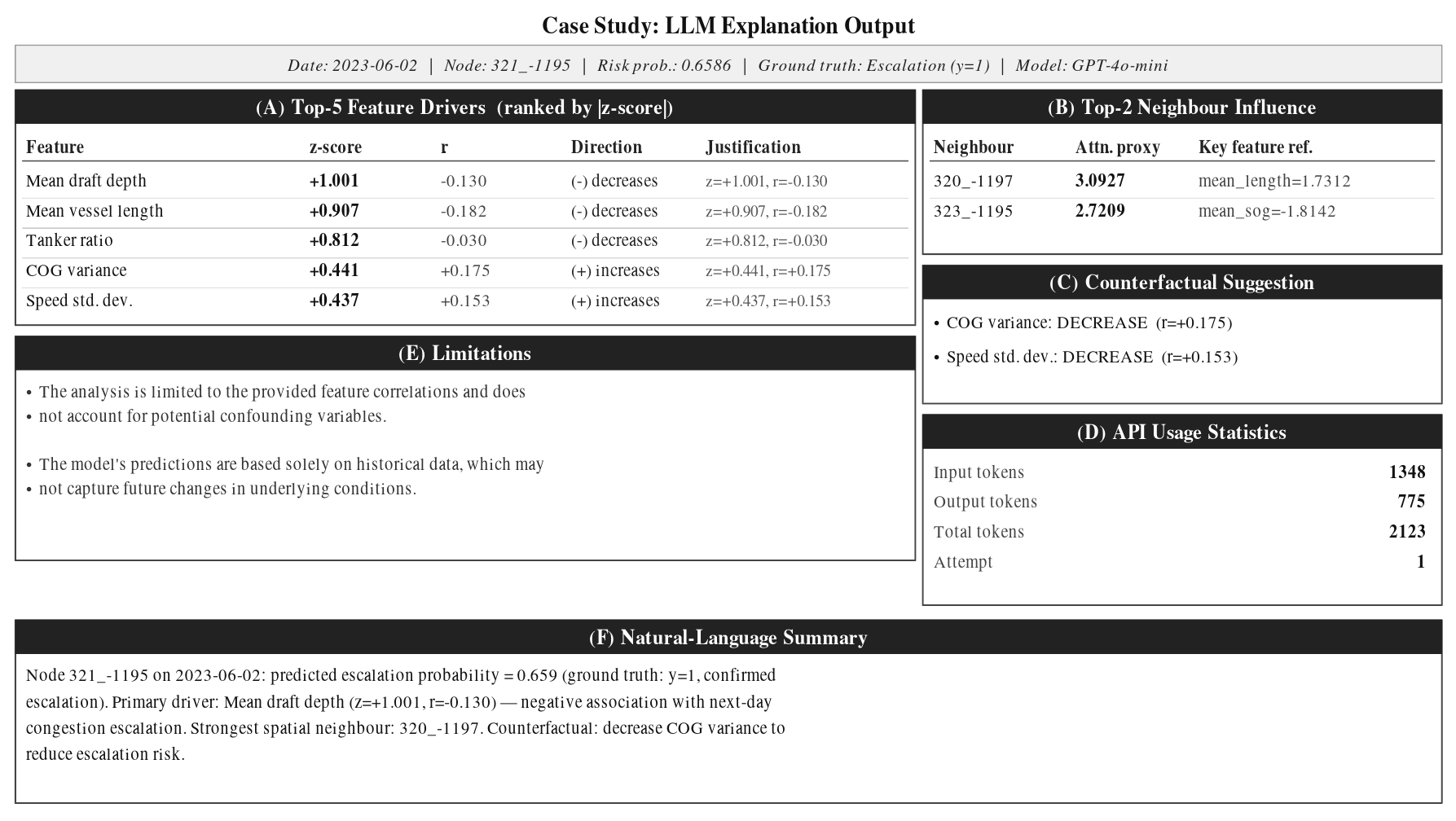}
  \caption{evidence record and generated risk report}
  \label{fig:system_architecture}
\end{figure*}
\subsection{Six-Section Report Structure}
The required output schema constrains the model to produce exactly six structured sections in JSON format: 
\begin{enumerate}
    \item \textbf{target feature drivers}, enumerating each of the top-5 features with a justification sentence linking the $z$-score and correlation to the predicted risk direction; 
    \item \textbf{neighbor influence}, describing how the attention weights and feature profiles of the top-2 spatial neighbors contribute to the node's predicted risk; 
    \item \textbf{risk summary}, a one-paragraph synthesis of the overall congestion outlook for that cell; 
    \item \textbf{counterfactual suggestions}, specifying minimal hypothetical feature adjustments that would change the prediction; 
    \item \textbf{confidence and uncertainty}, noting which features are ambiguous or have low correlation strength; and 
    \item \textbf{limitations}, explicitly acknowledging what the evidence record does not capture (e.g., exogenous events, weather, berth scheduling). 
\end{enumerate}
The JSON schema enforcement prevents free-form output and ensures that every section can be parsed programmatically and rendered in a human-readable dashboard.

\subsection{Directional-Consistency Validation}
To quantify the factual reliability of the generated reports, we define a \textit{directional-consistency check}: for each feature driver sentence in the generated output, we verify that the stated risk direction (\textit{increase risk} or \textit{decrease risk}) agrees with both the sign of the feature's $z$-score and the sign of its point-biserial correlation. A direction is consistent if the LLM's verbally expressed direction matches the label that was explicitly provided in the prompt. We apply this check across all 100 generated reports (100 node--day samples drawn from the test partition), evaluating each of the five feature driver entries per report, for a total of 500 direction judgments. 498 of the 500 judgments are consistent, yielding a directional-consistency rate of 99.6\%. The single inconsistent case arises from a \textit{tanker ratio} entry with a near-zero correlation (r = -0.017) where the model hedged its language in a way that was ambiguous to the parser. These results confirm that the generated explanations faithfully propagate the model-level evidence rather than fabricating plausible-sounding but unsupported risk attributions.

Fig.5 illustrates a representative case where node 321\_-1195 receives a congestion-escalation probability of 0.659. The explanation highlights slow ratio, std SOG, and two high-attention neighbors as the primary drivers of the predicted risk.

\section{Experimental Results}

\subsection{Quantitative Performance on the Test Set}

\begin{table}[htbp]
\caption{Test-Set Performance Comparison}
\begin{center}
\resizebox{\columnwidth}{!}{
    \begin{tabular}{|l|c|c|c|c|}
    \hline
    \textbf{Model} & \textbf{AUC} & \textbf{AP} & \textbf{F$_1$} & \textbf{Recall} \\
    \hline
    LR (no graph) & 0.713 & 0.300 & 0.375 & 0.480 \\
    \hline
    GCN (static graph) & 0.759 & 0.326 & 0.383 & 0.445 \\
    \hline
    \textbf{TGAT (graph + attention)} & \textbf{0.761} & \textbf{0.344} & \textbf{0.398} & \textbf{0.504} \\
    \hline
    \end{tabular}
}
\label{tab:performance_comparison}
\end{center}
\end{table}

Table I compares three models on the
binary supply chain risk-escalation task. Metrics are computed on the
held-out test partition (snapshots 76--89, 4531 node--day samples,
13.5\% positive rate). We report AUC, AP, $F_1$, and Recall;
Accuracy is omitted because a trivial all-negative classifier already
achieves 86.5\% accuracy on this imbalanced dataset, making it
uninformative.

LR establishes a non-graph linear baseline (AUC = 0.713). Although LR 
achieves a recall of 0.480---higher than GCN's 0.445---this reflects the 
behaviour of a class-balanced linear classifier that lowers its decision 
threshold aggressively to recover positives, at the cost of inferior AUC 
and AP. Adding static graph convolution (GCN) raises AUC to 0.759 but 
does not preserve this recall advantage, because sum-aggregation without 
learnable attention conflates high- and low-risk neighbours, producing 
less separable node embeddings for the minority class.
TGAT(graph with multi-head attention) achieves the best performance across all four metrics
(AUC=0.761, AP=0.344, $F_1\,=\,$0.398,
Recall$\,=\,$0.504). The recall gain over GCN (+13.2\%) is
operationally significant: for supply chain managers, missing a true
risk-escalation event carries a higher cost than issuing a
false alarm, making recall the primary deployment metric.

\subsection{Feature Importance Analysis}
To understand which features drive the model's predictions across the test set, we examine the distribution of $z$-score ranks from the attention evidence records. \textit{Mean SOG} and \textit{slow ratio} jointly appear in the top-2 feature positions for over $60\%$ of test-set nodes, consistent with their strong point-biserial correlations (r=-0.204 and r =+0.190, respectively). \textit{COG variance} (r = +0.181) and \textit{anchor ratio} (r = +0.176) frequently appear in positions three and four, while \textit{tanker ratio}, which has the weakest dataset-level correlation (r = -0.017), is rarely among the top-5 evidence features for high-confidence positive predictions.

The consistent alignment between the dataset-level correlation ranking and the model's per-node feature rankings provides an important sanity check: the TGAT is not over-relying on any single spurious feature, and the injected correlation directions in the LLM prompts are consistent with the model's implicit feature weighting.

\subsection{Discussion}
The marginal AUC gap between TGAT and GCN (0.761 vs 0.759) understates TGAT's advantage. On severely imbalanced datasets, Average Precision (AP) and Recall are more discriminative. TGAT achieves an AP of 0.344 (a +5.5\% gain over GCN) and a Recall of 0.504 (a +13.2\% gain). Recall is the primary deployment criterion for supply chain risk monitoring, where a missed risk escalation carries higher operational cost than a false alarm. These gains substantiate the operational benefit of attention-based spatiotemporal modeling.

The test AUC is achieved under a strict temporal split, preventing data leakage. The absolute recall of 0.50 is partly due to label noise inherent in the slow-ratio definition; future work using smoother label definitions (e.g., a 3-day moving-average) may improve this. GCN's lower recall confirms that static sum-aggregation cannot reliably distinguish high-risk nodes from low-risk neighbours---a limitation TGAT addresses by up-weighting spatially influential cells. Incorporating additional modalities, such as weather conditions or scheduled arrivals, represents a promising avenue for further gains.

\section{Conclusion}
This paper introduced an end-to-end framework that combines a Temporal Graph Attention Network with an LLM explanation module to predict and interpret localized bottleneck events within supply chain networks. Using six months of maritime logistics data as a representative case study for critical supply chain nodes, we achieved a test AUC of 0.761, AP of 0.344, and recall of 0.504 on a class-imbalanced benchmark without data leakage. We designed a structured LLM prompting pipeline that converts attention weights and $z$-score rankings into six-section natural-language early warning reports, validating the directional consistency at 99.6\%. To our knowledge, this is the first framework to combine graph-attention-based spatiotemporal prediction with LLM-generated, model-grounded explanations for supply chain risk early warning.

Promising extensions include expanding the framework to broader multimodal logistics networks (e.g., inland freight, rail), incorporating exogenous macroeconomic and environmental covariates to improve recall, and multi-step-ahead prediction via recursive temporal unrolling. Complementing $z$-score evidence with gradient-based SHAP~\cite{b5} or GNNExplainer~\cite{b4} attributions for multi-method corroboration is also planned. Deployment in a real-time supply chain risk management dashboard remains the ultimate validation target.

\end{document}